\documentclass[10pt,twocolumn,letterpaper]{article}

\usepackage{acpr}
\usepackage{times}
\usepackage{epsfig}
\usepackage{graphicx}
\usepackage{amsmath}
\usepackage{amssymb}
\usepackage{makecell}
\usepackage[numbers]{natbib}
\usepackage{lipsum}
\usepackage{comment}
\usepackage{lineno}
\usepackage{caption,subcaption}

\usepackage[pagebackref=true,breaklinks=true,letterpaper=true,colorlinks,bookmarks=false]{hyperref}

\acprfinalcopy 
\def\acprPaperID{124}
\ifacprfinal\fi
\begin{document}

\title{Dynamic Vision Sensors for Human Activity Recognition}

\author{Stefanie Anna Baby$^1$, Bimal Vinod$^2$, Chaitanya Chinni$^3$, Kaushik Mitra$^4$ \\
\and
Computational Imaging Lab\\
IIT Madras, Chennai, India\\
{\tt\footnotesize \{$^1$ee13b120,$^2$ee15m005,$^3$ee13b072,$^4$kmitra\}@ee.iitm.ac.in}
}

\maketitle


\begin{abstract} 
Unlike conventional cameras which capture video at a fixed frame rate, Dynamic Vision Sensors (DVS) record only changes in pixel intensity values. The output of DVS is simply a stream of discrete ON/OFF events based on the polarity of change in its pixel values. DVS has many attractive features such as low power consumption, high temporal resolution, high dynamic range and less storage requirements. All these make DVS a very promising camera for potential applications in wearable platforms where power consumption is a major concern.
In this paper we explore the feasibility of using DVS for Human Activity Recognition (HAR). 
We propose to use the various slices (such as $x-y$, $x-t$ and $y-t$) of the DVS video as a feature map for HAR  and denote them as Motion Maps. We show that fusing motion maps with Motion Boundary Histogram (MBH) gives good performance on the benchmark DVS dataset as well as on a real DVS gesture dataset collected by us. Interestingly, the performance of DVS is comparable to that of conventional videos although DVS captures only sparse motion information.  
%
%

\end{abstract}

\section{Introduction}
Conventional video camera uses frame based visual acquisition where each pixel is sampled at a fixed frame rate irrespective of whether or not their value changed. This leads to data redundancy and hence increased bandwidth and memory requirements. 
Dynamic Vision Sensor (DVS) \citep{R2} is a recent innovation in machine vision that mimics some of the functionalities of the human retinal vision. Instead of capturing the whole frame, it records only those pixels that see a change in intensity values. 
If the magnitude of change in log intensity value at a pixel is beyond a threshold an ON or OFF event is generated.

A major advantage of DVS is its ultra low power consumption. This is because it only generates ON/OFF events and avoids the use of ADCs which consumes the most power in conventional cameras. Hence DVS could be used to boost the battery life in  wearable or portable devices like untethered Augmented Reality (AR) devices, which currently use conventional cameras for various purposes such as gesture/activity recognition and building $3-$D maps. With this idea in mind, we explore performing activity/gesture recognition using DVS.

DVS is intrinsically suitable for gesture/activity recognition since it does not record any static information about the scene. Thus, we can avoid the overhead of  preprocessing algorithms such as background subtraction and contour extraction used in conventional image processing.
For the task of human activity recognition, we propose a simple method of using various  slices ($x-y$, $x-t$ and $y-t$) of the DVS video as feature maps. We denote these maps as \textit{motion maps} and employ Bag of Visual Words framework to extract critical features. Recognition rates obtained were similar to that of existing descriptors under this setting. We also combined the motion maps' features with state-of-the-art motion descriptor Motion Boundary Histogram (MBH) to obtain the best recognition rates, much higher than the HAR performance of individual descriptors. The results on DVS data are even comparable with the recognition rates seen in conventional videos. This is quite surprising given that DVS data is a very compressed version of the original video data with a remarkably sparse encoding. We experimented on two datasets: the DVS recordings of UCF11 \citep{hu2016dvs} and a hand gesture DVS dataset collected by us. In both the datasets our results have been promising for DVS.

\subsection{Related Work}


There are several works in literature for human activity recognition, of which we mention here a few relevant ones.
For a typical activity recognition task, two types of features are classically extracted - descriptors based on motion  and those based on shape. Motion History Images (MHI) from videos accumulate foreground regions of a person and accounts for its shape and stance \citep{bobick2001recognition}. Several more contour-based approaches such as Cartesian Coordinate Features, Fourier Descriptors Features  \citep{kauppinen1995experimental,de2000human}, Centroid-Distance Features and Chord-Length Features provide shape description  \citep{zhang2004review}. For motion based descriptors,  Histogram of Optical Flow (HOF) computes  optical flow of pixels between consecutive frames using brightness constancy assumption \citep{lucas1981iterative,bruhn2005lucas}.  Motion boundary histograms  take one step further by performing derivative operation on the optical flow field. This makes the feature invariant to local translation motion of the camera and captures only relative motion in the video \citep{dalal2006human, wang2013dense}.

Several other descriptors work by extracting the scene (background), color/hue and texture based features in a video. But texture and hue information is unavailable in DVS data because of its binary encoding scheme. The scene context based descriptors can also not be used with DVS videos since scenes usually are static in a video, unless there is significant camera motion. Nevertheless, volume based features like motion and shape  often provide sufficient information required to perform decent recognition and are more popular than surface features like color and texture.

Human activity recognition has been popularly solved by  extracting local features from videos on which Bag of Visual Words model (BoVW) is learnt and a classifier, typically SVM is trained \citep{yang2007evaluating, peng2016bag}.  As against this, recent works on HAR has focussed on deep learning techniques for improving recognition rates. Deep Convolutional and LSTM Recurrent Neural network units can be trained to automate feature extraction and directly perform natural sensor fusion \citep{ordonez2016deep} on human videos. Two-stream Convolutional Neural Networks learn the spatial and temporal information extracted from RGB and optical flow images of videos and are also becoming common for activity recognition \citep{ma2017ts, simonyan2014two}. However our method is simple and easy to implement, providing an intuitive framework for activity recognition.

Since DVS recording provide us with both motion and shape cues, we exploit these critical information by proposing a fusion of simple shape and motion based feature descriptors. 

\section{DVS Based Activity Recognition}

\begin{figure*}
\centering
\includegraphics[width=0.9\textwidth]{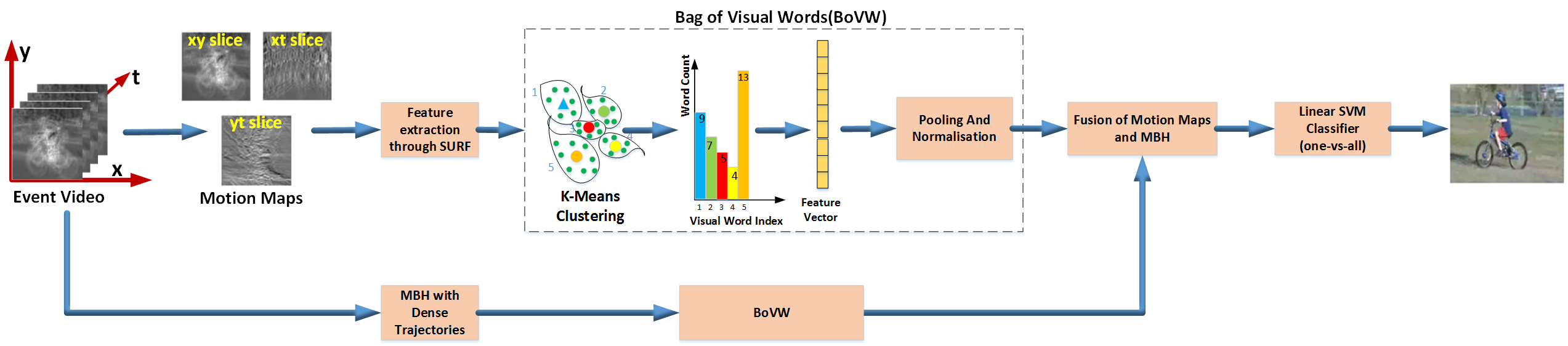}
\caption{ Our proposed method: The event stream from DVS is converted into video at $30 fps$. Motion maps are generated through various projections of this event video and SURF features are extracted. MBH features using dense trajectory are also extracted. Bag of features encoding from both these descriptors are combined and given to linear SVM classifier (\textit{one-vs-all}).}
\label{outline_pic}
\end{figure*}

Unlike conventional camera, DVS captures only motion and thus avoids the need to perform background subtraction. Also DVS output is usually sparse because change in pixel intensity occurs only at texture edges. To exploit this sparsity and motion cues captured by DVS, we propose to extract various projections of the DVS data (the motion maps) and use them as feature descriptor for activity recognition. Finally we fuse the motion maps  with a state-of-the-art motion descriptor MBH \cite{dalal2006human} to further improve the recognition accuracy. The overall architecture is shown in Figure \ref{outline_pic}.     

We first convert DVS event streams into a video by accumulating events over a time window. For our experiments we made videos at $30 fps$ framerate. From this, we obtain different $2$-D projections: $x-y$, $x-t$ and $y-t$ by averaging over each left-out dimension. Thus, $x-y$ projection is obtained by averaging over the $time$ axis, $x-t$ averages $y-axis$ and $y-t$ by averages $x-axis$. We call these $2$-D projections as motion maps since DVS captures the direction of motion of the foreground object. 

The $x-y$ motion map gives us the mean pose and stance of an object in the scene whereas the $x-t$ and the $y-t$ motion maps record the manner in which the object had moved over the video's duration. Our proposed $x-y$ motion map is similar to the idea of motion history images \citep{MHI} but we have two additional  maps that account for the movement of the object along the horizontal and vertical directions.

From the motion maps, we extract Bag of Features (BoF), where we use Speeded Up Robust Features (SURF) \citep{surf} extracted through grid search on the maps. 
This is followed by $k-means$ clustering of the train data’s features to create a visual vocabulary of $k$ words. Then features from each video are binned to these $k$ clusters and are $L_{2}$ normalized. Finally, a \textit{linear} SVM classifier under \textit{one-vs-all} encoding scheme is trained on the encoded features to predict the performed activity. Since the motion maps inherently complement each another with the $x-y$ map encoding the shape and pose of the object while the $x-t$ and the $y-t$ motion maps describing its motion, we combine all the three motion maps' descriptors to obtain better classification accuracy. We tried fusion of features before as well as after performing BoF and observed that fusion after BoF performs better. This result is in line with that of  \cite{peng2016bag} where the authors give a comprehensive study of fusing feature descriptors at several stages of BoF.


For the final recognition task, the motion maps descriptors are further combined with the MBH descriptor since MBH also encodes the appearance of objects and local motion in video but in a method that is distinctly different. MBH takes derivative of optical flow which in turn is computed using derivative of the video frames with respect to its spatial and temporal coordinates. Hence MBH employs second order statistics for its feature extraction while the motion maps use simple zero order statistics. Thus these two descriptors supplement one another and their combined feature set outperforms the individual  recognition rates. In the next section, we evaluate these descriptors and the loss in performance of HAR  on using DVS data compared to conventional videos on a benchmark dataset. We also report the performance on the DVS gesture dataset we have collected. From the results, we assess the usability of DVS for activity recognition and conclude with its shortcomings.

\section{Datasets}
We performed our experiments on two datasets - the UCF YouTube Action Data Set or UCF11 \citep{liu2009recognizing} and a DVS gesture dataset collected by us using DVS128.
 
The UCF11 data was chosen because it is one of the few human action datasets whose benchmark DVS counterpart is publicly available \citep{hu2016dvs}. The DVS data was created by the authors \cite{hu2016dvs} by re-recording the existing benchmark UCF11 videos played on a monitor using a DAViS240C vision sensor. Since the data was not directly recorded from the wild, this would mean that time resolution greater than that provided by the UCF11 video is not available in DVS under this simulated setting. Nonetheless, the dataset is sufficient for our experiments since it captures the sensor noise in DVS and is used on action videos that by themselves are not very fast paced.
 
\begin{figure}[!h]
\centering
\includegraphics[width=.42\textwidth]{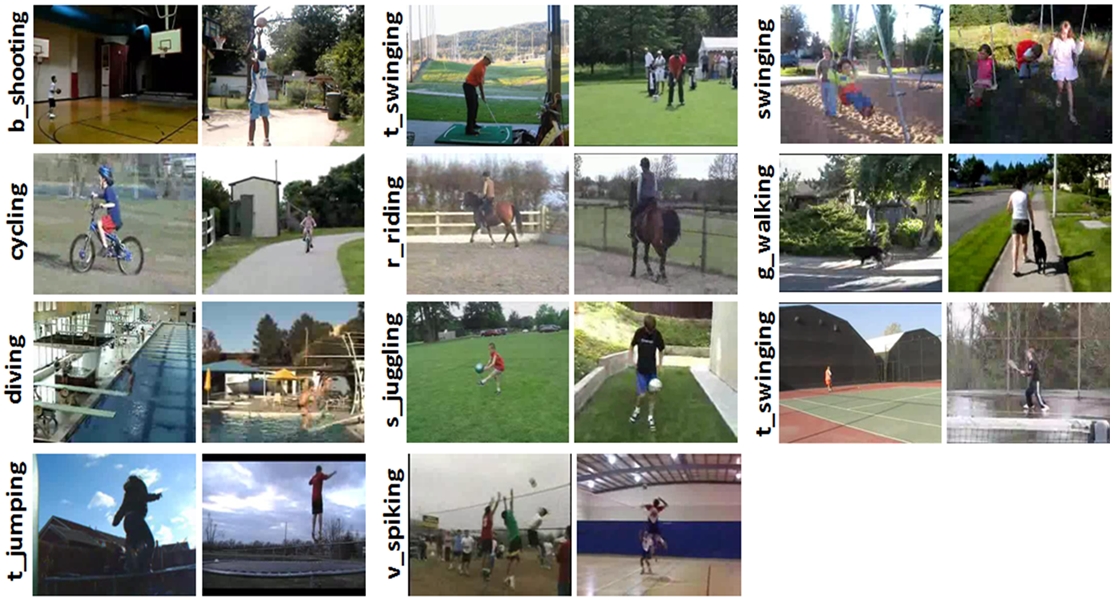}
\caption[The YouTube Action Data Set]{YouTube Action Data Set \footnotemark \\~ 
}
\label{UCF11}
\end{figure}

The UCF11 dataset contains eleven action classes as shown in Figure \ref{UCF11}, viz. basketball shooting, biking, diving, golf swinging, horse riding, soccer juggling, swinging, tennis swinging, trampoline jumping, volleyball spiking and walking dog.
Each class is further subdivided in to $25$ groups that allow us to perform Leave One Out (LOO) cross validation twenty five times on the actions as suggested by the creators of the data.

\footnotetext{Image source: \url{ http://crcv.ucf.edu/data/UCF_YouTube_Action.php}}

\begin{figure}[!h]
\centering
\includegraphics[width=.36\textwidth]{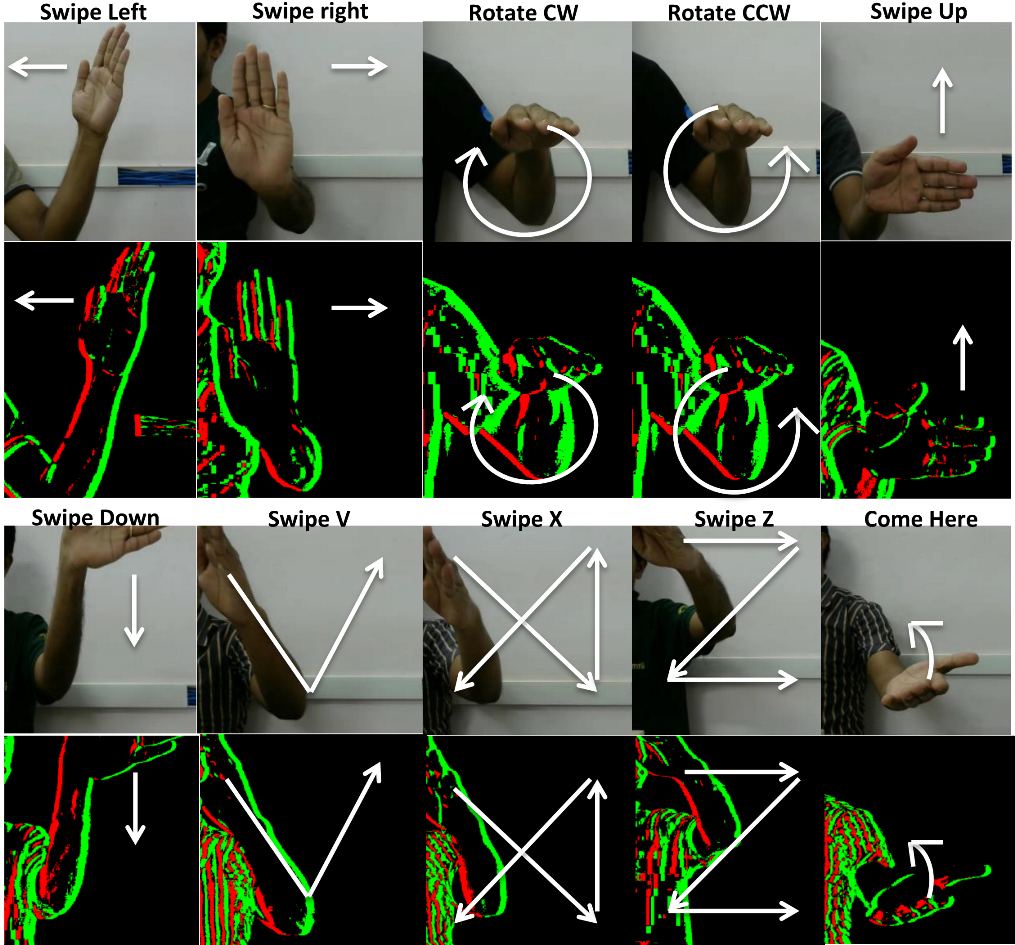}
\caption{Gestures from the DVS dataset collected by us. Ground truth from an RGB camera is also shown. 
}
\label{DVS_dataset}
\end{figure}
For a second round of experiments, we used the DVS hand gesture dataset collected by us using DVS128 and we refer them as the \textit{DVS Gesture data} (see Figure \ref{DVS_dataset}). The dataset contains $10$ different hand gestures, each performed $10$ times by $12$ subjects constituting a total of $1200$ gestures. The hand gestures used are left swipe, right swipe, beckon, counter-clock wise rotation, clock wise rotation, swipe down, swipe up, swipe V, wave X and wave Z. We performed $12$-fold cross-validation for all experiments on this dataset leaving out one subject each time. 

\begin{figure*}[ht]
  \centering
  \begin{subfigure}[b]{0.33\linewidth}
    \centering\includegraphics[width=120pt]{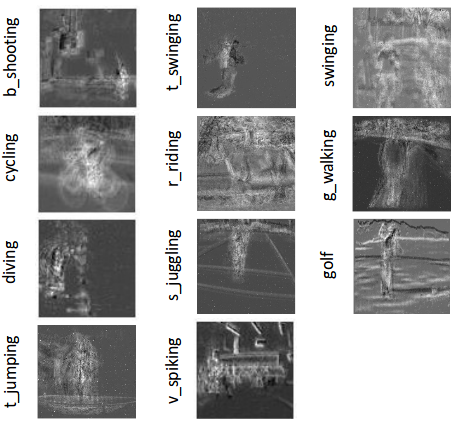}
      \label{fig:fig1}
    \caption{$x-y$ Motion Map}
  \end{subfigure}%
  \begin{subfigure}[b]{0.33\linewidth}
    \centering\includegraphics[width=120pt]{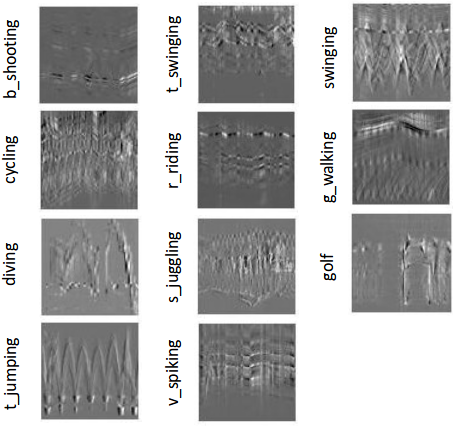}
    \label{fig:fig2}
    \caption{$x-t$ Motion Map}
  \end{subfigure}
  \begin{subfigure}[b]{0.33\linewidth}
    \centering\includegraphics[width=120pt]{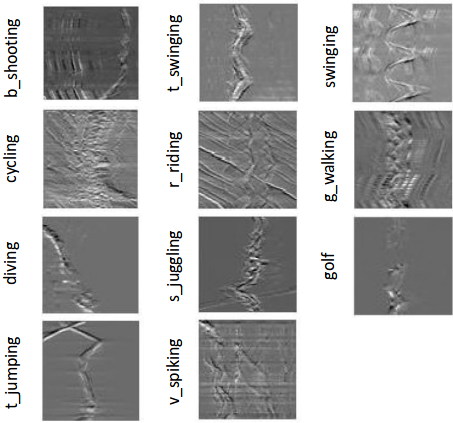}
    \label{fig:fig3}
    \caption{$y-t$ Motion Map}
  \end{subfigure}
  \caption{
  The three Motion maps for $11$ randomly picked UCF11 videos.}
    \label{fig_motion}
\end{figure*}

Figure \ref{fig_motion} shows the motion maps created from randomly picked videos of the eleven classes of UCF11 data. Note that in the $x-y$ map, much of the shape and pose of the object is captured. Similarly, the $x-t$ and $y-t$ slices show rhythmic patterns based on the movement involved typical for a given action category. Notable ones among these are winding river-like $y-t$ motion map for action class \textit{swinging} and the rhythmic up and down spikes in the $x-t$ motion map for \textit{trampoline} class.

\section{Feature Extraction and classification}
This section describes the steps that we used for feature extraction and classification. To evaluate existing motion descriptors like HoG, HOF and MBH, we extracted local spatio-temporal features using dense trajectories from the videos  \citep{wang:2011:inria-00583818:1}. Dense trajectories are created by dense sampling of images frame wise. The sampled points are tracked along frames using dense optical flow field and the trajectory length is limited to $15$ frames to avoid drifting of tracked points. Along the path tracked, descriptors like HoG, HOF or MBH are computed using the neighborhood volume of $32 \times 32 \times 15$ pixels. This volume is further divided into cells of size $16 \times 16$ pixels $\times 5$ frames. So each tracked tube gives a $2 \times 2 \times 3$ cells. Within each cell, the histograms of descriptors are found. For HoG and MBH, we used $8$ orientation bins per cell and the magnitude of the feature values were used for weighting. For HOF, an additional bin was added to account for pixels whose flow value was smaller than a threshold. All the descriptors were also $L_{2}$ normalized before performing bag of features. In total, HoG gave feature descriptors of size $96$ per tracked volume ($2 \times 2 \times 3$ cells per tracked path times $8$ bins) while HOF produced $108$ features ($2 \times 2 \times 3$ cells times $9$ bins). MBH also gave $96$ features similar to HoG, but in both horizontal and vertical directions. Thus, overall it had twice the number of features for a chosen trajectory. Bag of features was individually performed on each of these descriptors. Since each video produced about $\approx 500,000$ dense tracks, most of them in close proximity to one another, BoF was done on a subset of training features on $100,000$ trajectories randomly selected. To ensure that every video in the train set contributes to the codebook, we selected features randomly from each video instead of  pooling all extracted features first and performing random selection. The codebook dimension in the clustering step was maintained at $500$. After learning the cluster centers, all features of the video were used to generate the histograms of the same $500$ bins. Finally the segregated features were $L_{2}$ normalized and SVM classifier was trained.

On each motion map also we individually performed bag of features  with a codebook of dimension $500$. We have used Matlab's built-in function \texttt{bagOfFeatures} for this step and trained \textit{one-vs-all linear} SVM for the multi-class recognition. The results under Leave One Out cross-validation method for all these descriptors are given in the next section.

\section{Experimental Results}
We have conducted our experiments on two datasets as explained in the following section\footnote{For our code and DVS gesture dataset refer -  \url{https://github.com/Computational-Imaging-Lab-IITM/HAR-DVS}}.
\subsection{HAR on UCF11 and its DVS counterpart}
In this experiment, HAR was performed on the original UCF11  dataset (RGB) and its corresponding DVS recordings. Table \ref{results:ucf11} provides the recognition rates obtained with 25 fold Leave One Out cross-validation  method.

\begin{table*}[htbp]
	\centering
    \noindent
    \begin{subtable}{\textwidth}
    	\noindent
        \centering
        \scalebox{0.84}{%
         \begin{tabular}[c]{|c|c|c|c|c|c|c|c|c|c|} \hline
         Dataset & HoG & HOF & MBH & \thead{\textit{x-y}\\Motion Map} &\thead{\textit{x-t}\\Motion Map}& \thead{\textit{y-t}\\Motion Map} & {\thead{Combined\\ Motion Maps}}& {\thead{Motion Maps \\+ HOF}} & {\thead{Motion Maps \\+ MBH}}\\ \hline
         Original UCF11 & 0.6319 &	0.5754 &	0.7707 & 0.4397 & 0.4567 & 0.4077 & 	0.5867 & 0.6922 &	\textbf{0.7933}\\ \hline
          DVS recordings of UCF11 & 0.5358 &	0.6043 &	0.7016 & 0.4943 &	0.451 & 0.4629	& 0.6727 & 0.7299 & 	\textbf{0.7513}\\ \hline
        \end{tabular}}
        \caption{Results on UCF11 and its DVS counterpart}
        \label{results:ucf11}
    \end{subtable}%

    \begin{subtable}{\textwidth}
        \centering
         \begin{tabular}[c]{|c|c|c|c|c|c|c|c|c|c|}
			\multicolumn{10}{c}{\vspace{0.05cm}}\\
		\end{tabular}
    \end{subtable}%

    \begin{subtable}{\textwidth}
        \centering
        \scalebox{0.88}{%
         \begin{tabular}[c]{|c|c|c|c|c|c|c|c|c|c|} \hline
         Dataset & HoG & HOF & MBH & \thead{\textit{x-y}\\Motion Map} &\thead{\textit{x-t}\\Motion Map}& \thead{\textit{y-t}\\Motion Map} & {\thead{Combined\\ Motion Maps}}& {\thead{Motion Maps \\+ HOF}} & {\thead{Motion Maps \\+ MBH}}\\ \hline
         DVS gesture dataset& 0.8768 & 0.9689 & 0.9468 & 0.7748 & 0.8349 & 0.7899 & 0.9529 & 0.9809 & \textbf{0.9880} \\ \hline
        \end{tabular}}
        \caption{Results on the DVS gesture dataset collected by us}
        \label{results:dvs}
    \end{subtable}%

  \label{tab:sample}
  \caption{Recognition rates for various motion and shape descriptors on the UCF11 dataset, its corresponding DVS data and the DVS gesture dataset collected by us. Note that MBH features give $70\%$ accuracy while addition of motion maps give $75\%$ accuracy on the DVS recordings of UCF11 data}
       \label{table:results_01}
\end{table*}

\begin{figure}[!h]
\centering
\includegraphics[width=.41\textwidth]{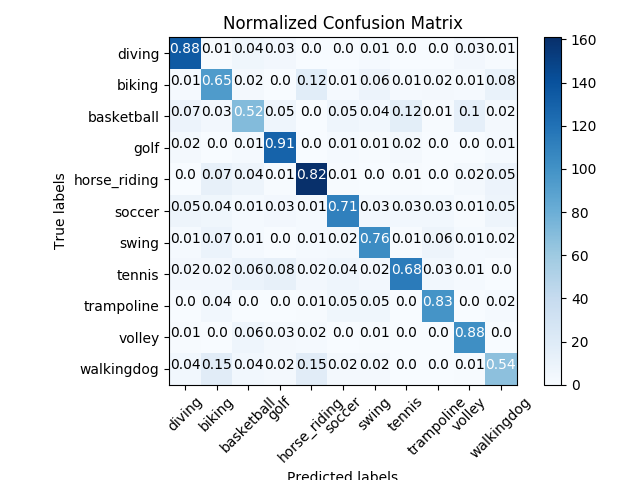}
\caption{Confusion matrix for UCF11-DVS dataset on combining motion maps and MBH}
\label{conf_matrix}
\end{figure}


The results show that fusion of motion maps from the DVS data gave a HAR rate of $67.27\%$,  comparable to the rates for HOF and HoG on the original UCF11 data. Interestingly, with the individual motion map descriptors the DVS recordings of UCF11 gave higher recognition rates while descriptors like HoG and HOF performed better on the original videos. This is because there is no background clutter and scene information in DVS recording for distracting its bag of features encoding. KNN classifier was also used for the final predictions, but it gave consistently about $5 \%$ lower HAR rates. Similarly, it was observed that simply using a larger dimension codebook of size $4000$ improves recognition rates by $2-3 \%$. Because our aim is to study the performance of DVS data for HAR compared to original data, we limited our codebook size to 500 words since using higher sized codebook simply improved both the results.

To further boost HAR rates, we separately included the MBH and HOF descriptors along with the motion maps and trained SVM classifier in light of the complementarity they offer. The HAR values in Table \ref{results:ucf11} show that the features from motion maps better complement the second order statistics of MBH than the
first order HOF features on both the UCF11 datasets. The results also show that the fusion of MBH and motion maps gave the highest recognition rate among all descriptors and nearly bridged the performance gap between DVS and conventional videos on the benchmark UCF11 data. Given the sparsity of DVS data, it is remarkable that the final descriptor has provided a near-equivalent performance on DVS when compared to the results on conventional videos.  

\subsection{Recognition on our DVS gesture dataset}


\begin{figure*}[!ht]
  \centering
  \begin{subfigure}[b]{0.49\linewidth}
    \centering\includegraphics[width=\textwidth]{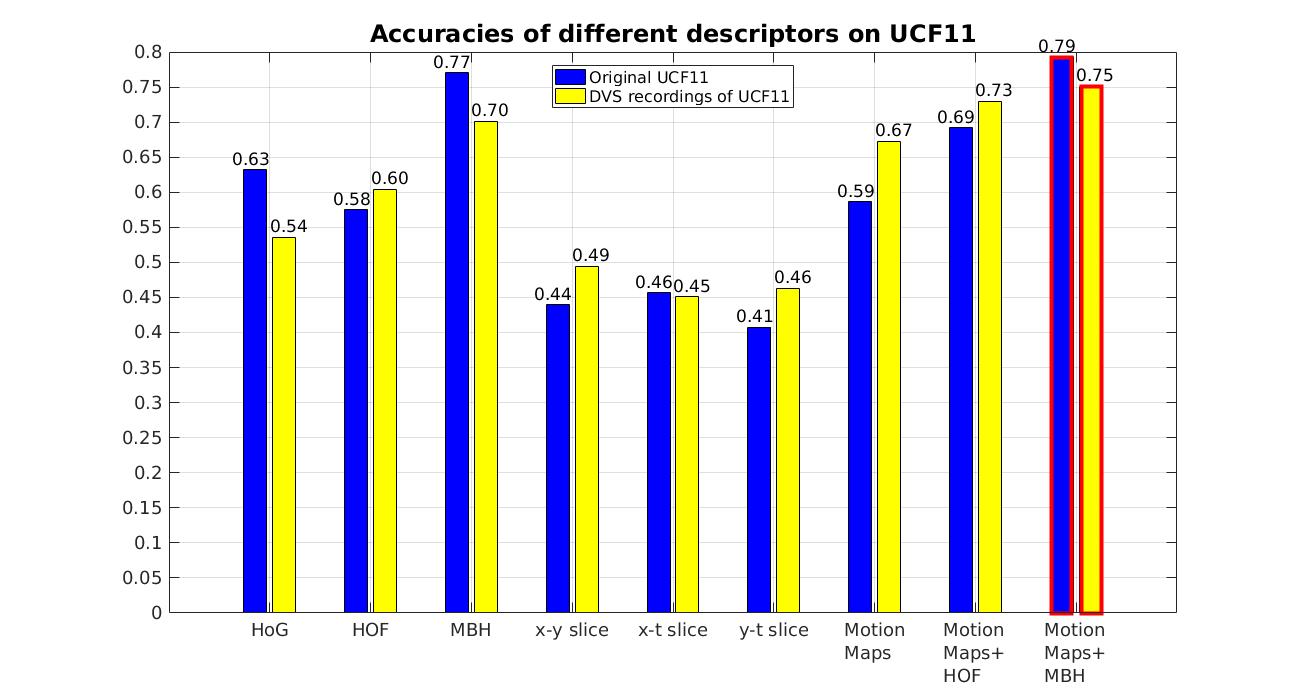}
      \label{fig:accuracy_ucf11}
    \caption{Accuracies on original UCF11 and its DVS counterpart}
  \end{subfigure}%
  \begin{subfigure}[b]{0.49\linewidth}
    \centering\includegraphics[width=\textwidth]{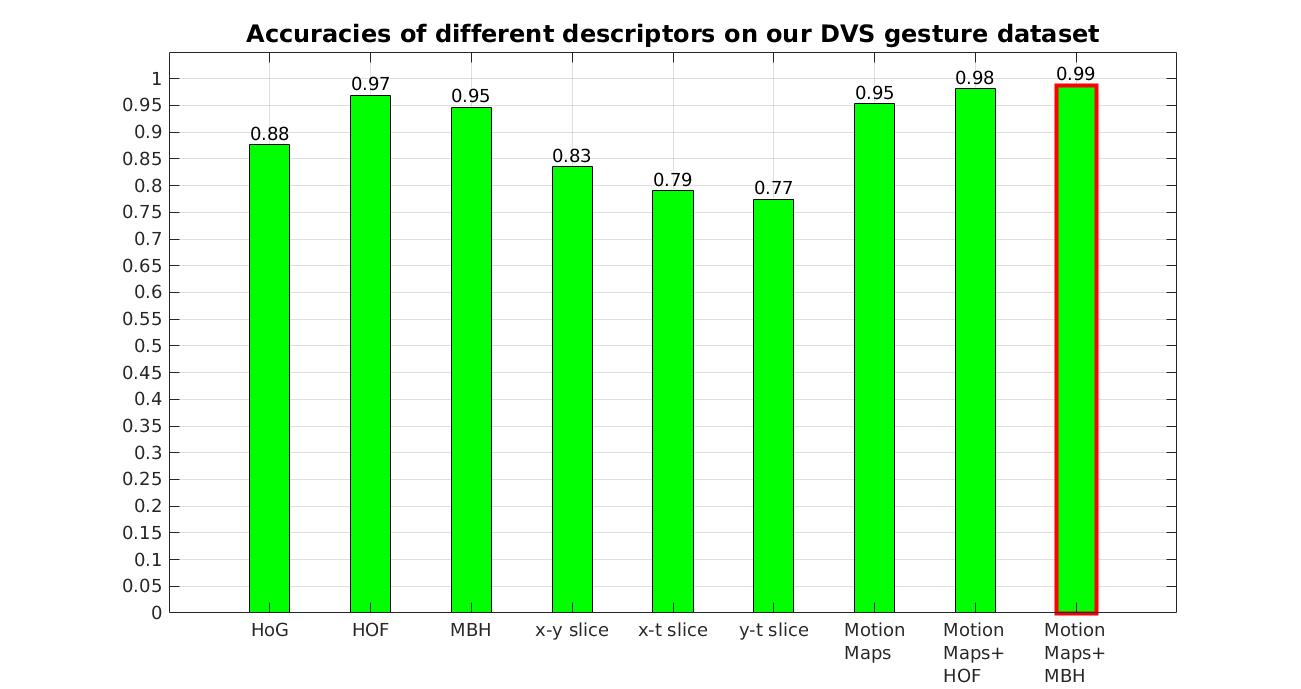}
    \label{fig:acc_iitm}
    \caption{Accuracies on our DVS gesture dataset}
  \end{subfigure}
  
  \caption{
  Accuracy plots on UCF11 and our DVS gesture dataset.
    \label{accuracy_plot}
  }
\end{figure*}

With our  DVS gesture dataset also, we obtained decent recognition rates by combining the motion maps alone, nearly same as that given by existing feature descriptors. Combining motion maps with the MBH descriptor again augmented HAR performance to  give the highest recognition rates as seen in Table \ref{results:dvs}. 

\section{Conclusion and Future Work}
In this paper, we have analyzed the performance of DVS data in human activity recognition and compared it with its conventional frame-based counterpart using traditional feature extraction techniques. We also proposed a new encoding technique (motion maps) that is suited especially for DVS data in light of its sparse and concise recording scheme. Combining the existing MBH descriptor with motion maps gave the best recognition results.

Based on the feature descriptors  available for its encoding, HAR results from DVS recordings have been nearly equal to that of RGB videos on the benchmark UCF11 data.  Additional features based on the scene, texture and hue have enabled better recognition rates with actual videos. But these are more complex and unavailable for use with the DVS data from the very beginning. Hence respecting the limitations that come with DVS, we conclude that within the framework of its possible descriptors it is just as useful for HAR as conventional videos and could efficiently be used in place of the latter, especially in low power and high speed applications.

As future work, we can look at improving performance of simple bag-of-features where location based relations are not preserved due to its pooling step. Rather than destroying spatial information between the extracted features in the image,  methods like \textit{Spatial Correlogram} and matching can be employed on the DVS data. Also, we noted that similar to recognition rates in conventional videos, dense trajectories with MBH gave the best results on using traditional features in DVS as well. Much of the success of dense tracking comes from the fact that it generates too many interest points given any video sample. Visualization of the  interest points found by dense sampling showed that some of these are randomly fired noisy events in DVS unrelated to the object in foreground. A simple median filtering pre-processing before finding dense interest points however \textit{did not} improve recognition rate. In order to truly address the problem, a new method specifically for finding and tracking DVS events should itself be invented. This would act as the true initial step for improving the performance of HAR on using optical flow, MBH as well as dense trajectories with dynamic vision sensors.


{\small
\bibliographystyle{ieee}
\bibliography{dvshar}
}

\end{document}